# KorQuAD1.0: Korean QA Dataset for Machine Reading Comprehension

Seungyoung Lim, Myungji Kim, Jooyoul Lee
LG CNS, AI Bigdata Research Center

**Abstract**

Machine Reading Comprehension (MRC) is a task that requires machine to understand natural language and answer questions by reading a document. It is the core of automatic response technology such as chatbots and automatized customer supporting systems. We present Korean Question Answering Dataset(KorQuAD), a large-scale Korean dataset for extractive machine reading comprehension task. It consists of 70,000+ human generated question-answer pairs on Korean Wikipedia articles. We release KorQuAD1.0 and launch a challenge at https://KorQuAD.github.io to encourage the development of multilingual natural language processing research.

## 1. Introduction

Standard datasets play an essential role for algorithmic research in that recent deep neural networks require vast datasets to learn high level features and perform the task. They also provide a criterion for performance evaluation between several models. Nonetheless, there are very few Korean data published as a standard for natural language processing, unlike English. Researchers have to either rely on translators to translate English standard data or construct the dataset themselves at cost, which is not only burdensome but difficult to compare performance between multiple architectures.

Due to the growing demand and lack of standard dataset in Korean, we introduce KorQuAD1.0 (Korean Question Answering Dataset), a large-scale question-and-answer dataset constructed for Korean machine reading comprehension. This dataset benchmarks the data generating process of SQuAD1.0[1] to meet the standard. The dataset is freely available at KorQuAD website.

We launch KorQuAD1.0 as a challenge so that researchers can evaluate their models with the official criterion. For better understanding of the dataset, the statistics of investigated distribution of answers and the types of reasoning required to answer the question are included in the paper.

We contribute to multilingual language processing research through the introduction of KorQuAD1.0, helping researchers by providing a large amount of learning data for question answering tasks based on machine reading and evaluate the performance of research achievements on an objective basis.

## 2. Existing Datasets

The desideratum for KorQuAD1.0 is that it should function as a standard dataset for machine reading comprehension task. We thus survey existing standard machine reading comprehension datasets for English. SQuAD (Stanford Question Answering Dataset) is the representative extractive reading comprehension dataset consisting of 100,000+ questions. Recently, 50,000 unanswerable questions were supplemented to this dataset and released as SQuAD2.0[2] so that machine must decide whether the question is answerable. Hotpot QA[3] is another extractive reading comprehension dataset where the main feature is that it requires multi-hop reasoning over multiple paragraphs. MS Marco(Microsoft Machine Reading Comprehension Dataset)[4] is answer-generating reading comprehension task that deals with multiple paragraphs for Bing web queries.

Aforementioned datasets are standard data for MRC task, and each of them runs a challenge and leaderboard, which serves as the stimulus to encourage research.

For Korean reading comprehension, K-QuAD[5] consists of 70,000 questions that are translations of SQuADv1.1 paragraph and question pairs plus 4,000 hand-crafted questions. The significant difference of the KorQuAD1.0 from this dataset is that all of the KorQuAD1.0 questions are human generated for well-formed Korean Wikipedia articles. We exclude any form of automatized translation to meet our goal of creating complex and lexically diverse natural language questions.

## 3. Dataset Collection

We benchmark the data collecting process of SQuAD1.0 and crowdsourced 70,000+ question-answer pairs. Most of data generating steps resemble that of the exemplary dataset, but we suggest a specific task guide created under the distinct characteristics of the Korean question answering task for rich vocabulary usage and various syntax.

### 3.1 Collecting Passages

We select documents for KorQuAD1.0 from Korean Wikipedia articles. In Wikipedia, 100 and 43 articles are designated as 'Alchan-geul' and 'Joeun-geul' [1] respectively, which meet the criterion of good content with proper configuration. To retrieve high-quality articles, we first collected all articles in these lists, and then additionally obtained 1,494 articles through random sampling. In total, 1,647 articles are collected. Next, we extracted individual paragraphs and removed images, tables, and URLs. Finally, we discarded passages that are either shorter than 300 characters or containing mathematical functions. 1,420 articles are used for the training set, 1,420 for the development set, and 140 for the test set.

### 3.2 Collecting Question-answer Pairs

Next, we employed crowdworkers and collected six questions for each passage, on average. One worker created a maximum of three questions per each passage, and we allocated one passage to 3 workers so that lexical and syntactic diversities can be imposed naturally. We guided workers to read through the given passage, and then type in the question in natural language. Crowdworkers were encouraged to make questions in their own word, ask with various forms of question such as 'casual' or 'respectful', create hard questions requiring reasoning, and refrain from asking questions about simple translations. We asked workers to select a minimal answer span that answers the question in the morpheme unit. In total, 70,079 question-answer pairs were collected and Table 1 shows statistics for the number of examples in KorQuAD1.0.

[1] https://ko.wikipedia.org/wiki/위키백과:알찬_글과_좋은_글의_차이점

|  | # Articles | #Paragraphs | #Questions |
|---|---|---|---|
| **Train Set** | 1,420 | 9,681 | 60,407 |
| **Dev Set** | 140 | 964 | 5,774 |
| **Test Set** | 77 | 623 | 3,898 |

**Table 1**: Statistics of dataset in KorQuAD1.0

## 4. Dataset Analysis

To understand the properties of questions and answers in KorQuAD1.0 dataset, we sample two questions from 140 paragraphs in the development set to conduct qualitative analysis on them. As a result, we conclude that KorQuAD1.0 requires systems to be robust to lexical and syntactic variations in natural language and to learn to infer the object, person, time, place, process, and reason to answer the question.

### 4.1 Type of Questions

In Table 2, we define six categories of reasoning required to answer the question and suggest the result of manual inspection for 280 questions sampled from the development set.

The most frequently asked question type accounts for 56.4% of the questions, which query either by changing the order of the wording of the supporting sentence or by reorganizing the syntax. Questions expressed with different vocabularies from passage using a synonym and world knowledge account for 13.6% and 3.9% respectively. Questions requiring the collection of evidence from multiple sentences account for 19.6%. 3.6% of the questions included a deduction for the choices in the sentence that meet the conditions of the question, or the higher level of reasoning using the information in parentheses. Finally, it is found that 2.9% of the questions are asked using the external knowledge that was not in the paragraph or the answer area is incorrectly selected due to the error of the worker.

| Type 1. Syntactic variation (56.4%) |
|---|
| Q: What is the novel that the writer Kim Young-ha won the 1st New Writer's Award given by Munhakdongne in 1996? |
| In 1995, he started his career as a writer by publishing the short-story <A Meditation On Mirror> in the quarterly magazine <Review>, and in 1996, he won the 1st Munhakdongne New Writer Award with the novel <I Have a Right to Destroy Myself>. |

| Type 2. Lexical variation - synonymy (13.6%) |
|---|
| Q: What did Sutherland create which is considered to be the **origin** of Augmented Reality research? |
| Augmented Reality research has **started** with the development of see-through HMD by Ivan Sutherland … |
| **Type 3. Lexical variation - world knowledge (3.9%)** |
| Q: What is the name of a project group formed with Kayip, Superdrive residing **overseas**? |
| He formed the project group 'mo:tet' with Kayip residing in **London, UK** and Superdrive residing in **Berlin, Germany** … |
| **Type 4. Multiple sentences reasoning (19.6%)** |
| Q: Why did Clemens have no chance of earning access to the Hall of Fame? |
| **All of these pitchers, except Clemens, are in the Hall of Fame**. Only Clemens was denied entry … It is uncertain whether he would be inducted into the Hall <u>since he is involved in the use of performance-enhancing drugs.</u> |
| **Type 5. Logical reasoning requirement (3.6%)** |
| Q. Who is related to Ministry of Justice among the ones failed to be nominated as candidates for primary election during the 17[th] presidential election in South Korea? |
| Former Uri party chairmen Chung Dong-young, … , **former Minister of Justice <u>Chun Jung-bae</u>**, registered for primary election. … **Chung Dong-young, Sohn Hak-kyu, Lee Hae-chan, Han Myeong-sook and Rhyu Si-min were nominated as candidates for primary election**. |
| **Type 6. Errors in questioning (2.9%)** |
| Q: What is the river originating from the Tibetan Plateau? |
| The rivers originate from the Tibetan Plateau, including Chang Jiang, <u>Huang He</u>, Indus, Tsangpo, … , Mekong, Irrawaddy and Salween. |

**Table 2**: Types of reasoning required to answer KorQuAD1.0 questions. Words relevant to the corresponding reasoning category are bolded and the ground truth answer is underlined. Examples with original Korean texts are attached in Appendix A.

### 4.2 Type of answers analysis

| Object | Person | Date | Location | Method | Reason |
|---|---|---|---|---|---|
| 55.4% | 23.2% | 8.9% | 7.5% | 4.3% | 0.7% |

**Table 3**: Answer types of KorQuAD1.0

We categorize the answers for the questions into six groups based on Table 3. As a result, object type answers account for 55.4%, followed by person, date, and location ones. Questions asking for method and reason account for 4.3% and 0.7% respectively. Compared to the answer types of SQuAD1.0 analyzed in [6], KorQuAD1.0 has a slightly higher proportion of object or person, but the portions of the other classes are similar.

## 5. Experiment & Results

As a baseline, we provide the performance of two end-to-end neural network models: S³-Net[7] and BERT[8]. S³-Net which exploits Korean morpheme level representation with sentence-level representation is known to perform well in Korean NLP tasks. BERT is a powerful pre-trained model that obtained state-of-the-art performance on various NLP tasks. We train both algorithms and provide their accuracy on KorQuAD1.0 test set. S³-Net is trained with the best hyper-parameters suggested in the original paper. We use the multi-lingual pre-trained model released by Google to fine-tune BERT model for KorQuAD1.0 task, without applying any Korean-specific natural language processing techniques.

We also employed additional workers to generate secondary answers on the test dataset. In this task, workers are asked to select the shortest span in the given passage that answers the given question. We compare the baseline model accuracy with that of human performance.

### 5.1 Experimental Results

We use two metrics to evaluate model accuracy.
**EM**: This metric measures the percentage of predictions that exactly match the ground truth answer.
**F1**: This metric measures the overlap between the prediction and the ground truth answer. Unlike SQuAD, we adopt character-level F1 that compute the percentage of overlapping characters, in that computing F1 based on the bags of tokens is not applicable for the Korean language due to various forms of grammar. Table 4 shows an example of computing F1 in KorQuAD1.0.

| Soon after the blessing, Nash quickly lowered the king's clothes under the roof. … put the clothes on and waited <u>for 5 days</u>(5일간). |
|---|
| Q: After the ritual, how many days did they wait after |

throwing the king's clothes and putting them directly on the dead king's body?

**Ground Truth**: 5일간  (*In English: for 5 days*)
**Predicted Answer**: 5일  (*In English: 5 days*)

| Token-based F1 | Char -based F1 | F1 for English |
|---|---|---|
| 0% | 80% | 80% |

**Table 4:** KorQuAD1.0 F1 computation example. In the Korean language, morphemes are used in various forms within the unit of spacing. Because there is no perfect morpheme analyzing system, it is impossible to calculate the F1 score based on the morpheme unit, yet the score based on the spacing unit does not reflect the introduction philosophy of F1. For this reason, we adopt character based F1 as our standard metric. Examples in original Korean texts are attached in Appendix A.

Table 5 shows the performance of baseline models alongside human performance. On the test set, S³-Net can return the answer span with EM 71.52%, F1 82.99%, which underperforms human. BERT outperforms S³-Net by 6.77%p in F1 score and 0.16%p in EM measure. The relatively subtle difference in EM score is due to ignorance of morpheme in the current model; the performance can be enhanced by adding Korean tokenizer so that the model can point out the exact span for the answer.

|  | Validation set |  | Test set |  |
|---|---|---|---|---|
|  | EM | F1 | EM | F1 |
| **S³-Net** | 71.68% | 82.87% | 71.52% | 82.99% |
| **BERT** | 70.89% | 90.24% | 71.68% | 89.76% |
| **Human** | - | - | 80.17% | 91.20% |

**Table 5**: Performance of the baseline model and human.

### 5.2 Result Analysis

To gain insight into the performance of the baseline models, we inspected the accuracy of S³-Net and BERT stratified by reasoning types of questions explored in Table 2. The results show that S³-Net is mostly challenged by lexical variation with a synonym or world knowledge: the model answers the question with only 60.53% of accuracy in the case of former, and in the latter case, the performance gap with the human is the biggest with 18%p. BERT performed better in lexical reasoning, suggesting the effectiveness of language pre-training. For queries requiring evidence aggregation or logical reasoning from several sentences, S³-Net performance is about 10%p less than that of the human. The model performance of BERT is worse for multiple sentences reasoning with 23%p gap with human performance, which need to be investigated further in future work. For the type of syntax variation, the gaps of both neural models are smaller than that of other types, although it still underperforms human.

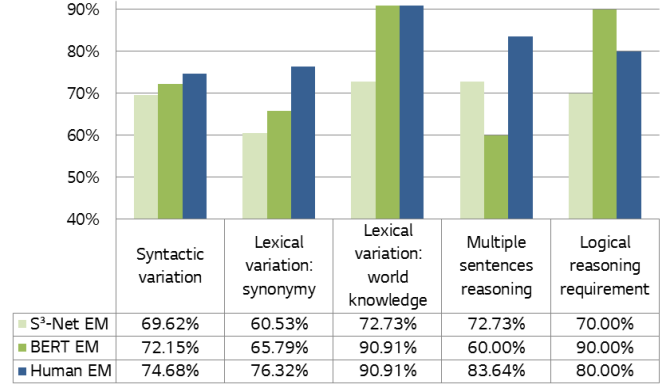

| | Syntactic variation | Lexical variation: synonymy | Lexical variation: world knowledge | Multiple sentences reasoning | Logical reasoning requirement |
|---|---|---|---|---|---|
| S³-Net EM | 69.62% | 60.53% | 72.73% | 72.73% | 70.00% |
| BERT EM | 72.15% | 65.79% | 90.91% | 60.00% | 90.00% |
| Human EM | 74.68% | 76.32% | 90.91% | 83.64% | 80.00% |

**Figure 1:** Results of S³-Net, BERT-multilingual and human performance stratified by type of questions. This result is based on 280 sampled dev questions inspected in Table 2.

### 6. Conclusion

We introduce KorQuAD1.0 - Korean Question Answering Dataset, a large-scale standard question-answering dataset and contribute to researchers in multilingual natural language processing. This data is collected on the same basis as the English standard data, SQuAD, and the properties of the data are similar. Accordingly, we present KorQuAD1.0 as standard data for Korean extractive machine reading comprehension task. The data is freely available through the GitHub site. We also launch KorQuAD1.0 as a challenge to encourage exploration of models and provide an evaluation of performance between models. We intend to continually create standard data regarding variety of QA research areas such as the task of assessing whether a system can know what it cannot answer, the task of extracting answers by among various documents, and the task of questioning on formatted documents that have structures such as tables[9] or web documents. We also will keep providing fair evaluation among models on standard datasets and contribute to the active multilingual language engineering research.

## Appendix

**A. Original Korean texts for examples**

| Type 1. Syntactic variation (56.4%) |
|---|
| Q: 김영하 소설가가 제 1회 문학동네 작가상을 수상한 작품으로, 96년 발표된 장편소설은 무엇인가? |
| 1995년 단편 <거울에 대한 명상>을 계간 《리뷰》에 발표하며 작품활동을 시작하였고 이듬해 96년 장편 《나는 나를 파괴할 권리가 있다》로 제 1회 문학동네 작가상을 수상하였다. |
| **Type 2. Lexical variation - synonymy (13.6%)** |
| Q: 증강현실은 서덜랜드가 무엇을 발전시킨 것을 시작으로 연구가 **시작**되었는가? |
| 이반 서덜랜드가 see-through HMD를 발전시킨 것을 **시초**로 하여 연구되기 시작한 증강현실은 … |
| **Type 3. Lexical variation - world knowledge (3.9%)** |
| Q: **해외**에서 활동하는 Kayip, Superdrive와 함께 결성한 프로젝트 그룹의 이름은? |
| **영국**에서 활동하고 있는 Kayip, **베를린**에서 활동하고 있는 Superdrive 와 함께 프로젝트 그룹 '모텟'을 결성 … |
| **Type 4. Multiple sentences reasoning (19.6%)** |
| Q: 클레멘스가 명예의 전당에 입성하지 못한 이유는? |
| … 이 투수들이 클레멘스를 제외하고 모두 명예의 전당에 올랐기 때문이다. 클레멘스만이 … 받았다. 그는 경기력 향상 약물 사용에 연루되어 있기 때문에 입성 여부가 불확실하다. |
| **Type 5. Logical reasoning requirement (3.6%)** |
| Q. 대한민국 제17대 대통령 선거 당시 후보로 등록했으나 예비경선의 경선 후보로 뽑히지 못한 사람 중 법무부와 관련 있는 사람은? |
| 정동영 전 열린우리당 의장, …, **천정배** 전 **법무부 장관**, … 등이 후보로 등록하였고… 예비경선으로 정동영, 손학규, 이해찬, 한명숙, 유시민 후보가 경선 후보로 결정되었다. |
| **Type 6. Errors in questioning (2.9%)** |
| Q. 티베트 고원에서 발원하는 강은? |
| … 강들이 티베트 고원에서 발원하는데, 창장, **황허**, 인더스, 사틀루즈, 창포 (…), 메콩, 이라와디, 살윈 강 등이 포함된다. |

**Table 6.** Examples of the types of reasoning required to answer KorQuAD1.0 questions in original Korean texts.

복을 하고 난 직후에 내시가 왕이 입고 있던 옷을 재빨리 지붕 아래로 … 그 옷을 덮고 5일간 살아나기를 기다렸다.

**Q**: 복의식 직후 왕의 옷을 아래에 있는 내시에게 던지면 곧장 죽은 왕의 몸 위에 덮고 며칠간을 기다렸는가?

**Ground Truth**: 5일간 (*In English: for 5 days*)
**Predicted Answer**: 5일 (*In English: 5 days*)

| Token-based F1 | Char-based F1 | F1 for English |
|---|---|---|
| 0% | 80% | 80% |

**Table 7.** Example of computing character based F1 in original Korean texts.